\documentclass[a4paper]{cas-sc}

\usepackage[numbers]{natbib}
\usepackage{array}
\newcolumntype{C}[1]{>{\centering\arraybackslash}p{#1}}

\begin{document}
\let\WriteBookmarks\relax
\def\floatpagepagefraction{1}
\def\textpagefraction{.001}
\shorttitle{On-Device Inference vs.\ Wireless Streaming for Wearable Cardiovascular Patches}
\shortauthors{M. F. R. Ibrahim et~al.}

\title [mode = title]{On-Device Inference versus Wireless Streaming: Energy-Efficient Multi-Modal Deep Learning for Wearable Cardiovascular Patches}                      



\author[1,4,5]{Mustafa Fuad Rifet Ibrahim}[orcid=0009-0008-3575-5688]
\cormark[1]
\ead{mustafa.ibrahim@nxp.com}

\credit{Conceptualization, Data curation, Formal analysis, Investigation, Methodology, Software, Validation, Visualization, Writing - Original Draft, Writing - Review \& Editing}

\affiliation[1]{organization={CTO System Innovation, NXP Semiconductors Germany GmbH},
                addressline={Beiersdorfstraße 12}, 
                city={Hamburg},
                postcode={22529},
                country={Germany}}

\author[2]{Tunc Alkanat}

\credit{Writing - Review \& Editing}

\affiliation[2]{organization={Advanced Chip Engineering, NXP Semiconductors},
	addressline={High Tech Campus 60}, 
	city={Eindhoven},
	postcode={5656},
	country={The Netherlands}}
	
\author[1]{Felix Manthey}

\credit{Software, Writing - Review \& Editing}

\author[3]{Maurice Meijer}

\credit{Writing - Review \& Editing, Supervision}

\affiliation[3]{organization={Business Line Secure Connected Edge, NXP Semiconductors},
	addressline={High Tech Campus 60}, 
	city={Eindhoven},
	postcode={5656},
	country={The Netherlands}}
	
\author[4]{Alexander Schlaefer}

\credit{Writing - Review \& Editing, Supervision}

\affiliation[4]{organization={Institute of Medical Technology and Intelligent Systems, Hamburg University of Technology},
	addressline={Am Schwarzenberg-Campus 1}, 
	city={Hamburg},
	postcode={21073},
	country={Germany}}

\author[5]{Peer Stelldinger}

\credit{Writing - Review \& Editing, Supervision}

\affiliation[5]{organization={Department of Informatics, Hamburg University of Applied Sciences},
	addressline={Berliner Tor 5}, 
	city={Hamburg},
	postcode={20099},
	country={Germany}}

\cortext[cor1]{Corresponding author}

\begin{abstract}
Wearable cardiovascular sensor patches promise continuous, unobtrusive monitoring, but their tight energy, memory, and compute budgets make it unclear whether physiological signals should be analyzed on the device or streamed to the cloud for processing. We study this inference-versus-transmission trade-off for a resource-constrained patch that records synchronized electrocardiogram (ECG) and phonocardiogram (PCG) signals. We propose an end-to-end, multi-modal convolutional neural network (CNN) with early fusion that classifies the two modalities directly on the device, without hand-crafted features. Trained and validated on the PhysioNet/Computing in Cardiology Challenge 2016 dataset, the floating-point model attains an accuracy of 0.975, which is competitive with the best reported results. At the same time, it reduces the parameter count and computational cost by approximately three orders of magnitude. We deploy an 8-bit integer version of the model on a microcontroller with an integrated neural processing unit (NPU) and measure its inference energy. We also benchmark the energy required for Bluetooth Low Energy (BLE) communication on a representative evaluation kit across a range of payload sizes. NPU inference consumes approximately one-seventh of the energy required for CPU inference. For realistic per-second payloads, local inference is also several times more energy efficient than continuous raw-data streaming. These results show that on-device intelligence, rather than constant transmission, is the more energy-efficient basis for always-on wearable cardiovascular monitoring at the edge.
\end{abstract}

\begin{keywords}
edge computing \sep wearable computing \sep on-device inference \sep TinyML \sep energy-efficient inference \sep mobile health \sep Bluetooth Low Energy
\end{keywords}

\maketitle

\section{Introduction}
\label{sec:introduction}
Pervasive health monitoring increasingly relies on small, body-worn devices that must sense, reason about, and communicate physiological data under severe energy, memory, and compute constraints. An important design question for wearable health-monitoring systems is whether data should be analyzed locally at the edge or transmitted to the cloud for processing. We investigate this trade-off for cardiovascular sensor patches. Cardiovascular diseases (CVDs) are the leading cause of death globally \cite{bhf2021}. Early detection of CVDs or associated risk factors significantly improves the efficacy of preventive measures \cite{stampfer2000primary, chiuve2006healthy}. Continuous monitoring is essential for early detection, and combining modalities such as electrocardiograms (ECGs) and phonocardiograms (PCGs) can enhance diagnostic performance \cite{li2021prediction}. To minimize the impact on daily activities, monitoring devices must be lightweight.

Continuous physiological monitoring generates large volumes of data and therefore requires automated processing. Deep learning~(DL) provides models that can analyze this data and has been shown to yield improved performance in the analysis of ECG and PCG data \cite{hong2020opportunities, chen2021deep}. However, these performance gains often require substantial computation and energy, which hinders deployment in portable medical applications. While cloud or hub devices (e.g., smartphones) can provide the computational power for DL, remote processing introduces privacy concerns, transmission energy costs, and potential service disruptions. We therefore focus on executing the DL model directly on the monitoring device. Performing the analysis locally on the device keeps the raw, continuously sampled signal on the patch and thereby enhances privacy. Transmission is required only when an anomaly is detected, which reduces energy consumption and allows for continuous and independent monitoring.

Computing on the sensor comes with very tight constraints on memory footprint, computational cost and energy consumption. Low energy consumption enables longer battery life, smaller batteries or even operation from a harvested power supply, all of which improve patient comfort. In this work we propose an efficient end-to-end convolutional neural network (CNN) \cite{lecun1989handwritten} built around a lightweight CNN building block. We evaluate its classification performance on a public dataset of synchronized ECG and PCG recordings. The model achieves accuracy competitive with the state-of-the-art while requiring approximately three orders of magnitude fewer parameters and FLOPs than the leading methods.

This work is a substantially extended version of our earlier paper~\cite{ibrahim2024end}, which approximated the hardware requirements of the proposed CNN through simulation. The present work turns that feasibility argument into a measured, systems-level result. Its central contribution is a direct, hardware-measured comparison of two operating strategies for a wearable patch: running the model on the device versus continuously streaming the raw signal over BLE. We also analyze the communication and compute energy budget that projects this comparison to continuous operation and to anomaly-triggered upload, establishing when on-device inference is the more energy-efficient choice. This work extends our earlier paper in six respects: (i) it replaces simulated estimates with direct energy and latency measurements on a Cortex-M33 CPU and an eIQ Neutron NPU; (ii) it adds the BLE communication-energy benchmark, the inference-versus-streaming analysis, and a communication and compute energy-budget projection for continuous monitoring; (iii) it compares the proposed CNN against MLP, LSTM, and transformer baselines at a matched parameter budget; (iv) it replaces the single random split with a 10-times-repeated 5-fold, recording-level cross-validation and an explicit discussion of the data limitations; (v) it adds a systematic input-length/downsampling study identifying the accuracy/energy operating point; and (vi) it expands the review of related work.

The remainder of this paper is organized as follows. Section~\ref{sect:rw} reviews related work. Section~\ref{sect:method} describes our method. Sections~\ref{sect:exp} and~\ref{sect:eff} present the experimental setup and results for, respectively, classification performance and on-device efficiency. Section~\ref{sect:disc} discusses these results, and Section~\ref{sect:conc} concludes.

\section{Related Work}
\label{sect:rw}

\subsection{ECG-/PCG-Based Anomaly Detection}
Various ML-based solutions for the classification of ECG and PCG signals have been proposed. Earlier approaches such as \cite{susivc2022identification} focused on traditional ML methods involving explicit segmentation of cardiac cycles and extensive feature engineering (time, frequency and statistical features), followed by classifiers such as logistic regression and support vector machines (SVMs) \cite{boser1992training}.

Recent studies have increasingly adopted DL-based approaches. A common strategy converts 1D signals into 2D representations to leverage architectures designed for image data. This includes converting PCG segments into intensity maps \cite{low2018automatic}, using continuous wavelet transforms (CWT) to generate scalograms \cite{hettiarachchi2021novel}, or applying techniques like the Gramian Angular Difference Field (GADF) \cite{qi2023residual}. Transfer learning has been employed to mitigate the scarcity of synchronized multi-modal data \cite{hettiarachchi2021novel}. Other DL approaches process signals directly in the time domain, often using dual-branch 1-D CNNs with feature fusion before the fully connected layers \cite{morshed2023deep}.

Hybrid approaches and ensembles have also been explored. These methods combine classic ML pipelines (expert-designed features) with end-to-end DL pipelines (raw signals or spectrograms) using a meta-learner \cite{gjoreski2020machine}.Alternatively, CNNs and long short-term memory (LSTM) networks \cite{hochreiter1997long} can extract features that are subsequently classified by an SVM \cite{li2021prediction}. More complex ensembles combine different architectures (e.g., BiLSTM on 1D data with GoogLeNet on 2D scalograms) using evidence theory for decision fusion \cite{li2022research}.

To better capture the interplay between the heart's electrical (ECG) and mechanical (PCG) activities, advanced feature fusion strategies have recently emerged. Architectures like CPDNet \cite{zhang2024co} and PACFNet \cite{li2025progressive} use three-branch designs (ECG-specific, PCG-specific, and fusion encoders) and employ sophisticated mechanisms such as progressive dense fusion \cite{zhang2024co} or progressive attention-based cross-modal fusion \cite{li2025progressive} to integrate features dynamically across multiple hierarchical levels. Toward end-to-end processing with minimal pre-processing, Thomae et al. \cite{thomae2016using} developed a small CNN-GRU model with attention using only the raw PCG signal as input, achieving an average of specificity and sensitivity of 0.55.

Our work differs from these approaches in three respects. First, it operates directly on raw, downsampled signals without computationally expensive feature extraction or signal transformation. Second, it achieves classification performance competitive with the state of the art. Third, its parameter count and computational cost are approximately three orders of magnitude lower than those of the leading architectures.

\subsection{Wearable and On-Device (Edge) Machine Learning for Cardiovascular Monitoring}
Recent literature has summarized developments in biosignal acquisition, telemonitoring, and the application of ML in wearable cardiovascular monitoring. Reviews have covered the use of wearables in heart failure management \cite{gautam2022artificial}, the integration of novel sensors and algorithms \cite{krittanawong2021integration}, and the accuracy of heart rate monitoring \cite{alugubelli2022wearable}. Several studies analyzed the use of AI and wearables for detecting CVDs and other conditions \cite{neri2023electrocardiogram, moshawrab2023smart, lee2022artificial}, highlighting the potential for early diagnosis while discussing associated software and hardware challenges.

Other works have implemented and optimized ML solutions for specific cardiovascular use-cases. Ribeiro et al. \cite{ribeiro2022ecg} analyzed the latency and energy efficiency of a quantized CNN for continuous ECG monitoring on mobile and energy-efficient CPUs (ARM Cortex-A55 and A53). Ran et al. \cite{ran2022homecare} employed pruning and quantization to accelerate CNN inference on an FPGA, validating their approach with real-time analysis. Lu et al. \cite{lu2021efficient} designed a custom hardware architecture featuring a fully pipelined processing unit array for high efficiency. Hsieh et al. \cite{hsieh2024ultra} explored ultra-low power solutions, proposing an analog implementation suitable for battery-less sensor nodes. In contrast, systems such as the one proposed by Jin et al. \cite{jin2009predicting, jin2009hearttogo} rely on transmitting wearable data to a smartphone for classification.

Our work differs from these contributions in two respects. First, we jointly process synchronized ECG and PCG signals. Second, we directly measure the energy consumption of both on-device inference and BLE communication using representative evaluation platforms.

\section{Methodology}
\label{sect:method}
\begin{figure*}
	\centering
	\includegraphics[width=1.0\textwidth]{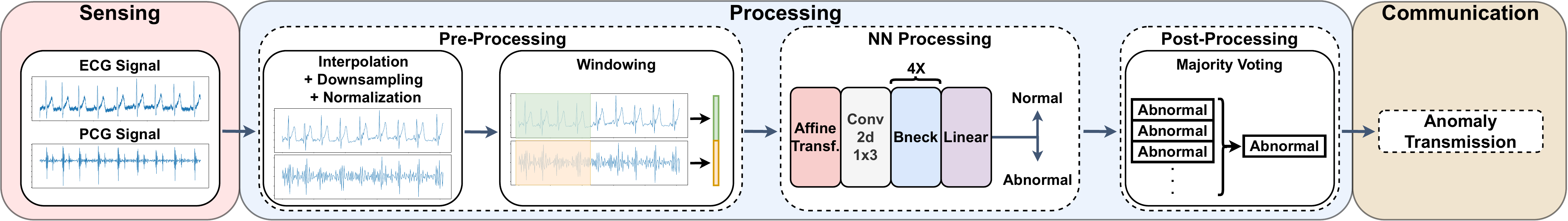}
	\caption{Overview of the system pipeline \cite{ibrahim2024end}. After acquisition, the ECG and PCG signals are interpolated, downsampled, and normalized before being divided into windows. Corresponding ECG and PCG windows are concatenated and classified as normal or abnormal by the proposed model. Majority voting across multiple windows produces the final classification. If an anomaly is detected, the associated data are transmitted to a clinician. (Reprinted with permission from M. F. R. Ibrahim et al., "End-to-End Multi-Modal Tiny-CNN for Cardiovascular Monitoring on Sensor Patches", 2024 IEEE International Conference on Pervasive Computing and Communications (PerCom),  979-8-3503-2603-1/24/\$31.00 \textcopyright2024 IEEE.)}
	\label{fig:system}
\end{figure*}

The system pipeline (Fig.~\ref{fig:system}) comprises three components: sensing (ECG and PCG), processing (classification) and communication on anomaly detection. We assume the use of a resource-constrained healthcare sensor patch. All of these operations incur computational cost, memory requirements (silicon area), and energy cost. The scope of this section is limited to the processing portion of the pipeline, since this is where our model is applied.

\subsection{Dataset}
\label{sect:method-A}
We use the PhysioNet/Computing in Cardiology Challenge 2016 dataset~\cite{challenge2016data, liu2016open, pollard2026physionet}. We use the Massachusetts Institute of Technology heart-sounds subset, designated “training-a,” because it provides synchronized ECG and PCG recordings. It is the only subset of the 2016 Challenge suitable for evaluating the complete multimodal pipeline. The remaining subsets (training-b through training-f) consist of PCG recordings only and therefore cannot support the multi-modal evaluation that is the focus of this study. The same constraint motivates our choice of the 2016 challenge rather than the 2022 challenge, which is dedicated to PCG-only murmur detection.

The "training-a" subset contains 409 recordings (405 of them with synchronized ECG and PCG) from 121 subjects. The size of the cohort and the recording count are properties of the original dataset rather than design choices on our part \cite{liu2016open}. The data were collected during in-home visits and hospital examinations. The class distribution is imbalanced, with 117 healthy and 288 pathological recordings. The pathologies represented are mitral valve prolapse, aortic disease, and benign murmurs.

Recordings range from 9 s to 37 s in length and were captured using a Welch Allyn Meditron electronic stethoscope and a single-lead ECG channel. The dataset documentation specifies the recording device but does not report the precise ECG lead configuration or electrode placement, which is a known limitation of this corpus. The data were downsampled from 44.1 kHz to 2 kHz when the challenge dataset was assembled by Liu et al. \cite{liu2016open}.

Crucially, the dataset lacks patient identifiers linking individual recordings back to specific subjects. This imposes limitations on the evaluation strategy that we discuss in Section~\ref{sect:exp} and revisit in Section~\ref{sect:disc}.

\subsection{Pre-Processing}
\label{sect:method-B}
We first interpolate any missing values in the ECG and PCG signals present in the dataset. We then significantly downsample the signals from 2 kHz to reduce computational cost and energy consumption on the target edge device. The signals are normalized separately to zero mean and unit variance, and 3 s windows with 2 s overlap are extracted. The window length is chosen to span at least one full cardiac cycle even at low heart rates, and the 2 s overlap supports the rolling majority vote in Section~\ref{sect:method-D}.

We aim for a per-modality tensor length suitable for efficient processing on a microcontroller, selecting a target length of 256 samples per modality based on the trade-off analysis presented in Sections~\ref{subsect:perfresults} and \ref{subsect:effresults}. From the 3 s window length, the target length determines the effective sampling frequency of 85.33 Hz. The two single-modality tensors are then concatenated along the width dimension to form an input of shape 1 x 1 x 512 (channels x height x width). The neural network operates end-to-end and requires neither heart-beat segmentation nor explicit feature extraction, which further reduces compute requirements.

We adopt early fusion through width-axis concatenation rather than channel stacking (i.e., shape 1 x 1 x 512 rather than 2 x 1 x 256). The two modalities are temporally synchronized but encode fundamentally different physical phenomena with different morphologies, frequency content, and signal-to-noise characteristics. Width-axis concatenation lets a per-modality affine layer (Section~\ref{subsect:networkdesign}) operate on each modality independently before any convolution and lets the convolution kernel learn cross-modal interactions through its receptive field rather than through fixed channel-wise mixing. The empirical effect of this choice, in combination with the affine layer, is reported in Table~\ref{table:Results2}.

In a practical, real-time deployment, the dataset interpolation step is unnecessary, provided that the sensor system delivers uniformly sampled data. Modern biopotential and acoustic acquisition front-ends typically use a hardware-triggered analog-to-digital converter and a fixed sample-rate clock, so the produced stream is uniformly sampled by construction. Interpolation is only required to handle exceptional dropouts. If the sensor produced non-uniform samples, an interpolation stage would be required before downsampling, but its energy cost would still be negligible compared with the convolutional pipeline. The same pre-processing is used for our proposed CNN and for the comparison architectures in Section~\ref{sect:exp}.

\subsection{Network Design}
\label{subsect:networkdesign}
\begin{table}[width=\columnwidth,cols=5,pos=h]
	\caption{Network Operations and Parameters}
	\label{table:NetworkDesign}
	
	\setlength{\tabcolsep}{3pt}
	
	\begin{tabular*}{\tblwidth}{@{}C{0.18\columnwidth}C{0.30\columnwidth}C{0.13\columnwidth}C{0.17\columnwidth}C{0.08\columnwidth}@{}}
		\toprule
		Input & Operator & Exp. size & Non-Lin. & Stride \\
		\midrule
		\(1\times1\times512\)  & Affine Transform          & -- & --        & -- \\
		\(1\times1\times512\)  & Conv2d, \(1\times5\)      & -- & HardSwish & 4  \\
		\(16\times1\times128\) & Bottleneck, \(1\times3\)  & 16 & ReLU      & 1  \\
		\(16\times1\times128\) & Bottleneck, \(1\times3\)  & 72 & ReLU      & 1  \\
		\(24\times1\times128\) & Bottleneck, \(1\times3\)  & 88 & ReLU      & 1  \\
		\(24\times1\times128\) & Bottleneck, \(1\times5\)  & 96 & HardSwish & 1  \\
		\(40\times1\times128\) & Avg. Pool                 & -- & --        & -- \\
		\(40\times1\times1\)   & Linear                    & -- & --        & -- \\
		\bottomrule
	\end{tabular*}
\end{table}

The proposed network has four parts. The operations and parameters of the primary configuration (CNN 256) are listed in Table~\ref{table:NetworkDesign}.

The first operation is a per-modality affine transform with separate, trainable scale and shift parameters for the ECG and PCG portions of the input. The two modality portions are equal in length (256 samples each in the primary configuration), but they differ substantially in physical scale, baseline distribution, and signal-to-noise characteristics. The affine layer therefore acts as a learnable, modality-specific normalization that lets the network up- or down-weight the contribution of each modality based on its task-relevant information content. The initial values are set with a coarse grid search. The empirical benefit of this layer is quantified in Table~\ref{table:Results2}.

The second part is a single convolution. For the primary CNN 256 configuration, this layer has kernel size 5 and stride 4, reducing the input length to 128 samples. When we vary the input signal length to study the accuracy/efficiency trade-off (Section~\ref{subsect:perfresults}), we adapt the kernel size and stride of this first convolution so that the output length remains constant at 128. This makes the receptive fields and the parameter budget of the downstream blocks directly comparable across configurations and isolates the effect of input rate.

The third part is a stack of four bottleneck blocks of the type introduced in MobileNetV2 \cite{sandler2018mobilenetv2}. These blocks keep the input/output dimensionality low while expanding it during the depthwise separable convolution, providing high expressiveness at low compute. We use the channel counts and non-linearities of the MobileNetV3-Small architecture \cite{howard2019searching}, replacing 2D with 1D convolutions and setting the stride to 1 in all bottleneck blocks to preserve temporal resolution after the first convolution.

The final part is a global average pooling followed by a fully connected layer with softmax activation that produces the two class scores. For comparison, we also implement a multilayer perceptron (MLP) \cite{rosenblatt1958perceptron}, an LSTM, and a transformer model \cite{vaswani2017attention} tuned for the same parameter budget. The detailed architecture parameters are listed in Appendix \ref{DLArch}. The same affine transform layer, with the same initialization, is used as the first operation in all three baselines.

\subsection{Post-Processing}
\label{sect:method-D}
To produce a recording-level classification, we apply a majority vote over all windows extracted from the recording. For real-time deployment, we propose a rolling majority vote over the current window and the previous N-1 windows. The parameter N governs a trade-off between detection latency and robustness to short-term fluctuations. Small N reacts quickly but is sensitive to single-window artifacts. Large N smooths the decision at the cost of slower response.

As a heuristic for choosing N, we recommend setting N x (window stride) to approximately the duration of the shortest clinical event of interest while keeping it short enough to satisfy the application's latency budget. With our 1 s effective stride, this yields N on the order of 5-10 windows for typical near-real-time monitoring. In our offline evaluation, recordings are 9-37 s long, which corresponds to roughly N = 7 to N = 35 windows when voting over a full recording. Per-deployment tuning on a target dataset is advisable, particularly when motion-artifact rates are high.

When an anomaly is detected, the data containing the anomaly is transmitted to a clinician for review. This requires buffering the N most recent windows, which is small in absolute terms thanks to the heavy downsampling. 

\section{Model Performance}
\label{sect:exp}

\subsection{Training}
We implement and train our CNN model in PyTorch \cite{paszke2019pytorch} on an NVIDIA RTX A6000. Training runs for 300 epochs using the AdamW optimizer \cite{loshchilov2017decoupled} and cross-entropy loss with an initial learning rate of \(10^{-3}\) and a batch size of 32. We address the class imbalance with weighted random sampling during training. To limit overfitting on the relatively small dataset, we rely on three mechanisms: the aforementioned weighted random sampling, the decoupled weight decay implicit in AdamW, and an extensive cross-validation protocol (Section~\ref{subsect:eval}). For the comparison architectures (MLP, LSTM, transformer), we use Optuna \cite{akiba2019optuna} to optimize hyperparameters, maximizing accuracy at a parameter count similar to our CNN model. The best configuration for each architecture is then trained with the same protocol as our CNN model.

\subsection{Evaluation}
\label{subsect:eval}
To robustly evaluate performance and generalization, particularly given the modest dataset size, we use 10 times repeated 5-fold cross-validation (CV). Ideally, cross-validation would be performed at the patient level. Recordings from the same patient may be strongly correlated, potentially inflating performance when they are assigned to both training and test folds. The training-a subset, however, lacks the per-recording patient identifiers needed for a patient-level split. We therefore split at the recording level, ensuring that all windows extracted from a single recording fall into either the training fold or the test fold of any given CV iteration. This prevents windows from the same recording from appearing in different folds. Such windows share the same sensor placement, local ambient noise, and short-term physiological state. Recording-level splitting is therefore the most rigorous protocol supported by the available metadata. The residual risk that two recordings from the same patient land on opposite sides of a split is discussed as a limitation in Section~\ref{subsec:VI-B}.

We report the average performance and the standard deviation across all 50 runs (10 repetitions × 5 folds). The standard deviation also serves as a measure of stability and sensitivity to the specific data partitioning. We measure performance with accuracy, sensitivity, specificity, precision, and F1 score \cite{blitzstein2019introduction}, plus the area under the receiver operating characteristic curve (AUC) for a threshold-independent assessment. We estimate memory cost via parameter count and computational cost via Floating-Point Operations (FLOPs).

For the comparisons with previous works in Table~\ref{table:Results1}, we estimated unreported efficiency metrics by reimplementing the architectures based on their published descriptions and analyzing them with the ptflops package \cite{ptflops}. When exact replication was infeasible, we estimated a lower bound by focusing on the computationally dominant portions (e.g., the CNN backbone). These estimates are intended to highlight order-of-magnitude differences in efficiency rather than to provide exact figures.

\subsection{Results}
\label{subsect:perfresults}
We evaluate our CNN by analyzing the impact of input signal length, comparing it with both state-of-the-art literature and other lightweight architectures, and ablating the input modalities.

\begin{table*}[width=\textwidth,cols=7,pos=h]
	\caption{Classification Results of Our CNN Model With Different Input Signal Lengths}
	\begin{tabular*}{\tblwidth}{@{}l@{\extracolsep{\fill}}cccccc@{}}
		\toprule
		\multicolumn{1}{c}{Model} & Accuracy                                                             & \multicolumn{1}{l}{Sensitivity}                                      & \multicolumn{1}{l}{Specificity}                                      & Precision                                                            & \multicolumn{1}{l}{F1 Score}                                         & \multicolumn{1}{l}{AUC}                                              \\ \midrule
		\textbf{CNN 6000}         & \textbf{\begin{tabular}[c]{@{}c@{}}0.9760\\ \(\pm\) 0.0227\end{tabular}} & \textbf{\begin{tabular}[c]{@{}c@{}}0.9854\\ \(\pm\) 0.0256\end{tabular}} & \begin{tabular}[c]{@{}c@{}}0.9667\\ \(\pm\) 0.0377\end{tabular} & \begin{tabular}[c]{@{}c@{}}0.9685\\ \(\pm\) 0.0344\end{tabular} & \textbf{\begin{tabular}[c]{@{}c@{}}0.9764\\ \(\pm\) 0.0222\end{tabular}} & \textbf{\begin{tabular}[c]{@{}c@{}}0.9770\\ \(\pm\) 0.0260\end{tabular}} \\ \midrule
		CNN 1024                  & \begin{tabular}[c]{@{}c@{}}0.9700\\ \(\pm\) 0.0252\end{tabular}          & \begin{tabular}[c]{@{}c@{}}0.9829\\ \(\pm\) 0.0246\end{tabular}          & \begin{tabular}[c]{@{}c@{}}0.9572\\ \(\pm\) 0.0466\end{tabular}          & \begin{tabular}[c]{@{}c@{}}0.9602\\ \(\pm\) 0.0412\end{tabular}          & \begin{tabular}[c]{@{}c@{}}0.9708\\ \(\pm\) 0.0242\end{tabular}          & \begin{tabular}[c]{@{}c@{}}0.9725\\ \(\pm\) 0.0302\end{tabular}          \\ \midrule
		CNN 512                  & \begin{tabular}[c]{@{}c@{}}0.9731\\ \(\pm\) 0.0214\end{tabular}          & \begin{tabular}[c]{@{}c@{}}0.9863\\ \(\pm\) 0.0237\end{tabular}          & \begin{tabular}[c]{@{}c@{}}0.9599\\ \(\pm\) 0.0397\end{tabular}          & \begin{tabular}[c]{@{}c@{}}0.9624\\ \(\pm\) 0.0361\end{tabular}          & \begin{tabular}[c]{@{}c@{}}0.9737\\ \(\pm\) 0.0207\end{tabular}          & \begin{tabular}[c]{@{}c@{}}0.9760\\ \(\pm\) 0.0259\end{tabular}          \\ \midrule
		CNN 256                  & \begin{tabular}[c]{@{}c@{}}0.9748\\ \(\pm\) 0.0227\end{tabular}          & \begin{tabular}[c]{@{}c@{}}0.9820\\ \(\pm\) 0.0246\end{tabular}          & \textbf{\begin{tabular}[c]{@{}c@{}}0.9675\\ \(\pm\) 0.0436\end{tabular}}          & \textbf{\begin{tabular}[c]{@{}c@{}}0.9698\\ \(\pm\) 0.0397\end{tabular}}          & \begin{tabular}[c]{@{}c@{}}0.9752\\ \(\pm\) 0.0218\end{tabular}          & \begin{tabular}[c]{@{}c@{}}0.9761\\ \(\pm\) 0.0271\end{tabular}          \\ \midrule
		CNN 128                   & \begin{tabular}[c]{@{}c@{}}0.9705\\ \(\pm\) 0.0246\end{tabular}          & \begin{tabular}[c]{@{}c@{}}0.9761\\ \(\pm\) 0.0376\end{tabular}          & \begin{tabular}[c]{@{}c@{}}0.9650\\ \(\pm\) 0.0402\end{tabular}          & \begin{tabular}[c]{@{}c@{}}0.9670\\ \(\pm\) 0.0363\end{tabular}          & \begin{tabular}[c]{@{}c@{}}0.9707\\ \(\pm\) 0.0248\end{tabular}          & \begin{tabular}[c]{@{}c@{}}0.9695\\ \(\pm\) 0.0349\end{tabular}          \\ \midrule
		CNN 64                    & \begin{tabular}[c]{@{}c@{}}0.9624\\ \(\pm\) 0.0257\end{tabular}          & \begin{tabular}[c]{@{}c@{}}0.9699\\ \(\pm\) 0.0305\end{tabular}          & \begin{tabular}[c]{@{}c@{}}0.9549\\ \(\pm\) 0.0457\end{tabular}          & \begin{tabular}[c]{@{}c@{}}0.9575\\ \(\pm\) 0.0416\end{tabular}          & \begin{tabular}[c]{@{}c@{}}0.9629\\ \(\pm\) 0.0250\end{tabular}          & \begin{tabular}[c]{@{}c@{}}0.9644\\ \(\pm\) 0.0330\end{tabular}          \\ \bottomrule
	\end{tabular*}
	\label{table:SigLengths}
\end{table*}

We tested input lengths from 6000 samples (the original 2 kHz at 3 s) down to 64 samples per modality (Table~\ref{table:SigLengths}). Across the full range, classification performance remains high: accuracy varies by less than 1.4 percentage points from CNN 6000 (0.9760) down to CNN 64 (0.9624), which is small relative to the \(\sim\)2.2 pp run-to-run standard deviation of each configuration. We therefore do not read the ordering among the input lengths as a meaningful ranking, and instead select the operating point on the energy axis. As reported in Section~\ref{sect:eff}, CNN 256 attains the lowest measured NPU energy per inference of the entire sweep (Table~\ref{table:ResultsExtended}), while its accuracy (0.9748) stays within run-to-run variation of the longest inputs. All subsequent results are reported for this CNN 256 configuration.

\begin{table*}[width=\textwidth,cols=9,pos=h]
	\caption{Reported classification performance and computational complexity of multimodal ECG-PCG models}
	\footnotesize
	\begin{tabular*}{\tblwidth}{@{}l@{\extracolsep{\fill}}cccccccc@{}}
		\toprule
		\multicolumn{1}{c}{Model}                           & Accuracy                                                    & \multicolumn{1}{l}{Sensitivity}                             & \multicolumn{1}{l}{Specificity}                             & \multicolumn{1}{l}{Precision}                               & \multicolumn{1}{l}{F1 Score}                                & \multicolumn{1}{l}{AUC}                                     & \multicolumn{1}{l}{Params} & \multicolumn{1}{l}{FLOPs} \\ \midrule
		W. P. Li, et al. (2025) \cite{li2025progressive}                         & \textbf{0.9777}                                              & 0.9799                                                        & \textbf{0.9728}                                              & \textbf{0.9879}                                              & \textbf{0.9839}                                              & \textbf{0.9967}                                                           & 11.7M                          & 3.14G                         \\ \midrule
		J. Li, et al. (2022)  \cite{li2022research}                        & 0.9613                                                      & \textbf{0.9848}                                             & 0.908                                                       & 0.9604                                                      & 0.9724                                                      & -                                                           & 5.8M                          & 3.0G                          \\ \midrule
		M. Morshed, et al. (2023) \cite{morshed2023deep} & 0.9506                                                      & 0.9506                                                      & 0.9092                                                      & 0.9508                                                      & 0.9500                                                      & 0.99                                               & 19.16M                          & 1.69G                        \\ \midrule
		H. Zhang, et al. (2024) \cite{zhang2024co}                        & \begin{tabular}[c]{@{}c@{}}0.9441\\ \(\pm\) 0.0304\end{tabular} & \begin{tabular}[c]{@{}c@{}}0.9485\\ \(\pm\) 0.0514\end{tabular} & \begin{tabular}[c]{@{}c@{}}0.9397\\ \(\pm\) 0.0497\end{tabular} & -                                                           & -                                                           & \begin{tabular}[c]{@{}c@{}}0.973\\ \(\pm\) 0.026\end{tabular}   & 25.3M                          & 3.84G                         \\ \midrule
		P. Qi, et al. (2023) \cite{qi2023residual}                        & 0.9434                                                      & 0.9769                                                      & 0.9092                                                      & -                                                           & -                                                           & -                                                           & 11.25M                          & 2.92G                         \\ \midrule
		P. Li, et al. (2021) \cite{li2021prediction}  & \begin{tabular}[c]{@{}c@{}}0.873\\ \(\pm\) 0.010\end{tabular}   & \begin{tabular}[c]{@{}c@{}}0.903\\ \(\pm\) 0.006\end{tabular}   & \begin{tabular}[c]{@{}c@{}}0.845\\ \(\pm\) 0.018\end{tabular}   & -                                                           & \begin{tabular}[c]{@{}c@{}}0.874\\ \(\pm\) 0.010\end{tabular}   & \begin{tabular}[c]{@{}c@{}}0.936\\ \(\pm\) 0.011\end{tabular}   & 446K                          & 59.1M                          \\ \midrule
		R. Hettiarachchi, et al. (2021) \cite{hettiarachchi2021novel}                & 0.9041                                                      & 0.9474                                                      & 0.75                                                        & -                                                           & -                                                           & 0.9106                                                      & 711K                         & 532M                       \\ \midrule
		\textbf{Our CNN}                  & \begin{tabular}[c]{@{}c@{}}0.9748\\ \(\pm\) 0.0227\end{tabular}          & \begin{tabular}[c]{@{}c@{}}0.9820\\ \(\pm\) 0.0246\end{tabular}          & \begin{tabular}[c]{@{}c@{}}0.9675\\ \(\pm\) 0.0436\end{tabular}          & \begin{tabular}[c]{@{}c@{}}0.9698\\ \(\pm\) 0.0397\end{tabular}          & \begin{tabular}[c]{@{}c@{}}0.9752\\ \(\pm\) 0.0218\end{tabular}          & \begin{tabular}[c]{@{}c@{}}0.9761\\ \(\pm\) 0.0271\end{tabular}         & \textbf{16K} & \textbf{3.74M}  \\ \bottomrule
	\end{tabular*}
	\label{table:Results1}
	\\[2pt]{\footnotesize Note: Reported classification metrics were obtained using heterogeneous evaluation protocols and are therefore not directly comparable. The table is intended to contextualize classification performance relative to parameter count and computational cost, rather than to establish a strict ranking.}
\end{table*}

Table~\ref{table:Results1} places our CNN against previously published DL models trained on the same dataset using both ECG and PCG signals. The reported metrics in the literature are obtained under heterogeneous protocols, some papers report accuracy at the segment level \cite{li2025progressive}, others use a single segment per recording \cite{li2022research}, and others evaluate at the recording level \cite{hettiarachchi2021novel, zhang2024co}. Because these evaluation protocols are not directly comparable, Table~\ref{table:Results1} should not be interpreted as a strict ranking of model performance. Instead, it provides context for the reported classification results relative to model size and computational cost. Within these limitations, CNN 256 achieves a recording-level accuracy of 0.9748, close to the highest value reported in the compared literature. Its principal advantage is the substantially lower parameter count and computational cost. Crucially, with only 16 K parameters and 3.74 M FLOPs, our model is approximately three orders of magnitude smaller than the leading entry (11.7 M parameters, 3.14 G FLOPs) and most other recent models \cite{zhang2024co, morshed2023deep, qi2023residual, li2022research}, while remaining over 27 times smaller than the smallest competing models \cite{li2021prediction, hettiarachchi2021novel}.

\begin{table*}[width=\textwidth,cols=9,pos=h]
	\caption{Classification Results Comparison With Other Network Architectures}
	\label{table:ResultsNAS}
	\footnotesize
	\begin{tabular*}{\tblwidth}{@{}l@{\extracolsep{\fill}}cccccccc@{}} 
		\toprule
		\multicolumn{1}{c}{Model} & Accuracy                                                                         & \multicolumn{1}{l}{Sensitivity}                                                  & \multicolumn{1}{l}{Specificity}                                                  & \multicolumn{1}{l}{Precision}                                                    & \multicolumn{1}{l}{F1 Score}                                                     & \multicolumn{1}{l}{AUC}                                                          & Params & FLOPs           \\ 
		\cmidrule{1-9}
		MLP                       & \begin{tabular}[c]{@{}c@{}}0.8290\\ \(\pm\) 0.0405\end{tabular}                   & \begin{tabular}[c]{@{}c@{}}0.8632\\ \(\pm\) 0.0842\end{tabular}                   & \begin{tabular}[c]{@{}c@{}}0.7948\\ \(\pm\) 0.0869\end{tabular}                   & \begin{tabular}[c]{@{}c@{}}0.8144\\ \(\pm\) 0.0620\end{tabular}                   & \begin{tabular}[c]{@{}c@{}}0.8338\\ \(\pm\) 0.0414\end{tabular}                   & \begin{tabular}[c]{@{}c@{}}0.8331\\ \(\pm\) 0.0555\end{tabular}                   & 16.32K & \textbf{31.9K}           \\ 
		\cmidrule{1-9}
		LSTM                      & \begin{tabular}[c]{@{}c@{}}0.8333\\ \(\pm\) 0.0485\end{tabular}                   & \begin{tabular}[c]{@{}c@{}}0.8683\\ \(\pm\) 0.0992\end{tabular}                   & \begin{tabular}[c]{@{}c@{}}0.7982\\ \(\pm\) 0.0997\end{tabular}                   & \begin{tabular}[c]{@{}c@{}}0.8206\\ \(\pm\) 0.0742\end{tabular}                   & \begin{tabular}[c]{@{}c@{}}0.8376\\ \(\pm\) 0.0517\end{tabular}                   & \begin{tabular}[c]{@{}c@{}}0.8410\\ \(\pm\) 0.0621\end{tabular}                   & 15.99K & 4.22M           \\ 
		\cmidrule{1-9}
		Transformer               & \begin{tabular}[c]{@{}c@{}}0.9282\\ \(\pm\) 0.0373\end{tabular}                   & \begin{tabular}[c]{@{}c@{}}0.9299\\ \(\pm\) 0.0666\end{tabular}                   & \begin{tabular}[c]{@{}c@{}}0.9265\\ \(\pm\) 0.0589\end{tabular}                   & \begin{tabular}[c]{@{}c@{}}0.9306\\ \(\pm\) 0.0520\end{tabular}                   & \begin{tabular}[c]{@{}c@{}}0.9278\\ \(\pm\) 0.0383\end{tabular}                   & \begin{tabular}[c]{@{}c@{}}0.9376\\ \(\pm\) 0.0361\end{tabular}                   & 16.04K & 2.46M           \\ 
		\cmidrule{1-9}
		\textbf{Our CNN}                  & \textbf{\begin{tabular}[c]{@{}c@{}}0.9748\\ \(\pm\) 0.0227\end{tabular}}          & \textbf{\begin{tabular}[c]{@{}c@{}}0.9820\\ \(\pm\) 0.0246\end{tabular}}          & \textbf{\begin{tabular}[c]{@{}c@{}}0.9675\\ \(\pm\) 0.0436\end{tabular}}          & \textbf{\begin{tabular}[c]{@{}c@{}}0.9698\\ \(\pm\) 0.0397\end{tabular}}          & \textbf{\begin{tabular}[c]{@{}c@{}}0.9752\\ \(\pm\) 0.0218\end{tabular}}          & \textbf{\begin{tabular}[c]{@{}c@{}}0.9761\\ \(\pm\) 0.0271\end{tabular}}         & 16K & 3.74M  \\
		\bottomrule
	\end{tabular*}
\end{table*}

Table~\ref{table:ResultsNAS} shows higher mean performance for the CNN than for the MLP, LSTM, and transformer baselines at a matched parameter budget. Possible reasons are discussed in Section~\ref{subsec:VI-A}.

\begin{table*}[width=\textwidth,cols=7,pos=h]
	\caption{Classification Results of Our CNN Model With Different Sensor Modalities}
	\begin{tabular*}{\tblwidth}{@{}l@{\extracolsep{\fill}}cccccc@{}} 
		\toprule
		\multicolumn{1}{c}{Model}                                             & Accuracy                                                             & \multicolumn{1}{l}{Sensitivity}                                      & \multicolumn{1}{l}{Specificity}                                      & \multicolumn{1}{l}{Precision}                                        & \multicolumn{1}{l}{F1 Score}                                         & \multicolumn{1}{l}{AUC}                                              \\ \midrule
		ECG Only                                                              & \begin{tabular}[c]{@{}c@{}}0.9496\\ \(\pm\) 0.0303\end{tabular}          & \begin{tabular}[c]{@{}c@{}}0.9521\\ \(\pm\) 0.0447\end{tabular}          & \begin{tabular}[c]{@{}c@{}}0.9471\\ \(\pm\) 0.0468\end{tabular}          & \begin{tabular}[c]{@{}c@{}}0.9495\\ \(\pm\) 0.0422\end{tabular}          & \begin{tabular}[c]{@{}c@{}}0.9497\\ \(\pm\) 0.0305\end{tabular}          & \begin{tabular}[c]{@{}c@{}}0.9546\\ \(\pm\) 0.0324\end{tabular}          \\ \midrule
		PCG Only                                                              & \begin{tabular}[c]{@{}c@{}}0.7854\\ \(\pm\) 0.0513\end{tabular}          & \begin{tabular}[c]{@{}c@{}}0.7789\\ \(\pm\) 0.1209\end{tabular}          & \begin{tabular}[c]{@{}c@{}}0.7919\\ \(\pm\) 0.1070\end{tabular}          & \begin{tabular}[c]{@{}c@{}}0.8011\\ \(\pm\) 0.0788\end{tabular}          & \begin{tabular}[c]{@{}c@{}}0.7806\\ \(\pm\) 0.0631\end{tabular}          & \begin{tabular}[c]{@{}c@{}}0.7605\\ \(\pm\) 0.0687\end{tabular}          \\ \midrule
		\begin{tabular}[c]{@{}l@{}}ECG + PCG\\ (no affine)\end{tabular}       & \begin{tabular}[c]{@{}c@{}}0.9509\\ \(\pm\) 0.0326\end{tabular}          & \begin{tabular}[c]{@{}c@{}}0.9743\\ \(\pm\) 0.0333\end{tabular}          & \begin{tabular}[c]{@{}c@{}}0.9274\\ \(\pm\) 0.0575\end{tabular}          & \begin{tabular}[c]{@{}c@{}}0.9333\\ \(\pm\) 0.0492\end{tabular}          & \begin{tabular}[c]{@{}c@{}}0.9524\\ \(\pm\) 0.0310\end{tabular}          & \begin{tabular}[c]{@{}c@{}}0.9582\\ \(\pm\) 0.0355\end{tabular}          \\ \midrule
		\textbf{\begin{tabular}[c]{@{}l@{}}ECG + PCG\\ (affine)\end{tabular}} & \textbf{\begin{tabular}[c]{@{}c@{}}0.9748\\ \(\pm\) 0.0227\end{tabular}}          & \textbf{\begin{tabular}[c]{@{}c@{}}0.9820\\ \(\pm\) 0.0246\end{tabular}}          & \textbf{\begin{tabular}[c]{@{}c@{}}0.9675\\ \(\pm\) 0.0436\end{tabular}}          & \textbf{\begin{tabular}[c]{@{}c@{}}0.9698\\ \(\pm\) 0.0397\end{tabular}}          & \textbf{\begin{tabular}[c]{@{}c@{}}0.9752\\ \(\pm\) 0.0218\end{tabular}}          & \textbf{\begin{tabular}[c]{@{}c@{}}0.9761\\ \(\pm\) 0.0271\end{tabular}} \\ \bottomrule
	\end{tabular*}
	\label{table:Results2}
\end{table*}

Table~\ref{table:Results2} quantifies the contribution of each sensor modality. The multi-modal model (ECG + PCG) with affine layer reaches 0.9748 accuracy versus 0.9496 for ECG only and 0.7854 for PCG only. Removing the affine transform from the multi-modal model drops accuracy to 0.9509, confirming that letting the network re-weight ECG against PCG is useful.

\begin{table*}[width=\textwidth,cols=7,pos=h]
	\caption{Classification Results Comparison Between Our 8-Bit Quantized and Non-Quantized CNN Model}
	\begin{tabular*}{\tblwidth}{@{}l@{\extracolsep{\fill}}cccccc@{}} 
		\toprule
		\multicolumn{1}{c}{Model} & Accuracy                                                             & \multicolumn{1}{l}{Sensitivity}                                      & \multicolumn{1}{l}{Specificity}                                      & \multicolumn{1}{l}{Precision}                                        & \multicolumn{1}{l}{F1 Score}                                         & \multicolumn{1}{l}{AUC}                                              \\ \midrule
		Not Quantized             & \textbf{\begin{tabular}[c]{@{}c@{}}0.9748\\ \(\pm\) 0.0227\end{tabular}}          & \textbf{\begin{tabular}[c]{@{}c@{}}0.9820\\ \(\pm\) 0.0246\end{tabular}}          & \textbf{\begin{tabular}[c]{@{}c@{}}0.9675\\ \(\pm\) 0.0436\end{tabular}}          & \textbf{\begin{tabular}[c]{@{}c@{}}0.9698\\ \(\pm\) 0.0397\end{tabular}}          & \textbf{\begin{tabular}[c]{@{}c@{}}0.9752\\ \(\pm\) 0.0218\end{tabular}}          & \textbf{\begin{tabular}[c]{@{}c@{}}0.9761\\ \(\pm\) 0.0271\end{tabular}} \\ \midrule
		Quantized                 & \begin{tabular}[c]{@{}c@{}}0.9055\\ \(\pm\) 0.0405\end{tabular}          & \begin{tabular}[c]{@{}c@{}}0.9144\\ \(\pm\) 0.0596\end{tabular}          & \begin{tabular}[c]{@{}c@{}}0.8966\\ \(\pm\) 0.0725\end{tabular}          & \begin{tabular}[c]{@{}c@{}}0.9033\\ \(\pm\) 0.0618\end{tabular}           & \begin{tabular}[c]{@{}c@{}}0.9065\\ \(\pm\) 0.0400\end{tabular}           & \begin{tabular}[c]{@{}c@{}}0.9118\\ \(\pm\) 0.0508\end{tabular}          \\ \bottomrule
	\end{tabular*}
	\label{table:Quant}
\end{table*}

For the on-device efficiency analysis in Section~\ref{sect:eff} we also quantize our CNN to 8-bit using quantization-aware training \cite{jacob2018quantization}. As shown in Table~\ref{table:Quant}, the quantized model attains 0.9055 accuracy compared with 0.9748 for the floating-point version. This precision/efficiency trade-off is discussed in Section~\ref{subsec:VI-A}.

\section{Model Efficiency}
\label{sect:eff}

\subsection{Setup}
\label{sect:eff-A}
We measure the energy consumption per inference and the inference time of all CNN variants on the MCXN947 evaluation kit (EVK) \cite{mcxnevk}. The EVK contains both an Arm Cortex-M33 CPU \cite{cortexm33} and the eIQ Neutron NPU \cite{bamberg2025eiq}, allowing a comparative analysis. The EVK is supplied at a board-input voltage of 3.3 V. However, the reported inference power is measured on the MCU CORE-domain rail rather than at the board input. Under the 150 MHz OverDrive configuration used in our experiments, the CORE-domain voltage is approximately 1.2 V. This configured voltage is used to convert the measured CORE-domain current into power.

We use the MCU-Link Pro Debug Probe \cite{mculink} to measure the mean power consumption of the CORE domain \cite{powermanagement}, which supplies the CPU and the NPU. Rather than reporting total board power, we report workload-specific incremental energy: the additional CORE-domain power drawn while the ML workload is running relative to an awake idle baseline on the same rail. For CPU inference, the baseline is measured with the NPU clock disabled and for NPU inference, it is measured with the NPU clock enabled. Consequently, the measured difference reflects the additional CORE-domain power associated with the respective inference workload.

The mean CORE-domain power is recorded both during inference and in the corresponding idle state. The difference between these measurements isolates the contribution of the inference workload. Energy per inference is then calculated from the incremental mean power during the inference window and the measured inference time. Because the model processes 3 s windows with an overlap of 2 s, a deployed pipeline performs one inference per second. At this rate, the numerical value of the energy per inference in millijoules equals the corresponding mean power contribution in milliwatts over one second. Each reported inference-time and energy value is the mean of repeated measurement runs. The corresponding standard deviations are given in Table~\ref{table:ResultsExtended}.

To compare on-device inference with continuous data streaming, we measure the energy consumption of BLE communication on the NHS52S04 EVK \cite{nhs52s04}. The EVK integrates the NHS52S04 ultra-low-power BLE SoC and includes an on-board MCU-Link current-measurement circuit. This circuit enables measurement of the isolated target-supply power of the NHS52S04 radio module (RDM), rather than the total EVK board power. The RDM supply path captures the combined MCU and RF activity of the NHS52S04 SoC. The NHS52S04 is operated in high-voltage supply mode, with the RDM supply path at ${\mathrm{VBAT\_HV}}\approx 3.0$ V. A supply voltage of 3.01 V, corresponding to the experimental configuration, is used to convert the measured RDM current into power.

As for inference, we report workload-specific incremental energy. The BLE increment is the additional power drawn through the RDM supply path while packets are transmitted over an established BLE connection, relative to a radio-off baseline in which the NHS52S04 CPU enters low-power sleep. It therefore captures the complete additional cost of enabling and maintaining the BLE communication path, including connection maintenance, protocol processing, CPU wake transitions, and RF activity. This measurement boundary reflects the two operating strategies considered in this work: during local inference, the radio can remain disabled in the absence of an anomaly, whereas continuous streaming requires the BLE communication path to remain active.

The inference and BLE increments are referenced to platform-appropriate, but not identical, baseline states. In particular, the BLE increment includes the small CPU wake-transition energy associated with the connection events. In a fully integrated implementation with a continuously active sensing processor, part of this activity might already be covered by the sensing workload. Connection maintenance and protocol processing energy, however, are genuine costs of the continuous-streaming strategy and are intentionally included because the radio can remain fully disabled during local inference.

To enable a direct comparison with the CNN variants, we sweep the transmitted payload $S_{\mathrm{data}}$ across six sizes: 128, 256, 512, 1024, 2048, and 12,000 B per second. These sizes span and extend beyond the data volume of a one-second ECG segment from typical consumer wearable ECG devices. For example, the Samsung Galaxy Watch samples ECG at 500 Hz \cite{SamsungWatch}, whereas the AliveCor KardiaMobile records at 300 Hz with 16-bit resolution \cite{AliveCor}. The comparison in Fig.~\ref{fig:PowerConsFig} therefore covers representative per-second payloads as well as a high-rate stress point. The relevant BLE link parameters are listed in Table~\ref{table:SensingComms}.

\begin{table}[width=\columnwidth,cols=2,pos=h]
	\caption{BLE Parameter Settings for Our Communication Experiment Setup}
	\label{table:SensingComms}
	
	\begin{tabular*}{\tblwidth}{@{}l@{\extracolsep{\fill}}c@{}}
		\toprule
		Parameter & \multicolumn{1}{l}{Value} \\ 
		\midrule
		BLE Tx power                       & 0 dBm                      \\
		BLE PHY mode                       & 2 Mbps                     \\
		BLE MTU                            & 247 Bytes                  \\
		BLE PDU                            & 251 Bytes                  \\
		Maximum payload, \(P_{\max}\)       & 244 Bytes                  \\
		BLE connection interval            & See Eq.~(\ref{eq:ci})      \\
		\bottomrule
	\end{tabular*}
\end{table}

Each per-second payload is divided into the minimum required number of approximately equal-sized packets, subject to the maximum payload
$P_{\max}$ of 244 B per packet. This minimizes the number of required connection events while distributing the transmitted data approximately evenly across them. The largest nominal connection interval that provides the required number of connection events within one second is calculated as

\begin{equation}
	\label{eq:ci}
	\mathrm{CI}
	=
	\left\lfloor
	\frac{1000\,\mathrm{ms}}
	{\left\lceil S_{\mathrm{data}} / P_{\max} \right\rceil}
	\right\rfloor .
\end{equation}

Here, $S_{\mathrm{data}}$ denotes the total data volume transmitted per second, $P_{\max}$ denotes the maximum payload carried by one packet, and $\left\lceil S_{\mathrm{data}}/P_{\max}\right\rceil$ is the minimum number of packets required to transmit the complete payload.

For payloads up to 2048 B/s, the configured connection intervals follow Eq. \eqref{eq:ci}, subject to the supported BLE connection-interval granularity. At 2048 B/s, for example, Eq. \eqref{eq:ci} gives a nominal interval of 111 ms, and the configured interval is 110 ms. At the highest tested rate of 12,000 B/s, however, the nominal 20 ms interval resulted in TX-buffer backpressure during sustained operation because it provided no surplus connection events. We therefore shortened the interval to 15 ms to provide scheduling margin and maintain stable transmission throughout the measurement.

To match the inference rate of one decision per second, the BLE configuration transmits each complete per-second payload within one second. Using the minimum required number of approximately equal-sized packets reduces protocol overhead and radio on-time while ensuring that the target data rate is sustained.

We fix the transmission budget at one payload per second for two reasons. First, this choice makes the communication cost directly comparable with the inference cost because the deployed inference pipeline likewise produces one decision per second. Second, and more fundamentally, the choice is approximately rate-invariant under continuous monitoring. If the link were operated less frequently, for example once every $u$ seconds, then each transmission would have to carry $u$ seconds of accumulated data to avoid gaps in the continuous record. The payload per transmission would therefore increase in proportion to the transmission interval. Batching changes the timing of the connection events but leaves the average transmitted data volume and associated radio activity approximately unchanged. Under a continuity constraint, transmitting larger payloads less frequently therefore does not eliminate the communication-energy cost; it primarily repackages the same data.

Sporadic monitoring that transmits only occasional short segments can reduce average communication energy, but at the cost of potentially missing intermittent events. This is precisely what continuous cardiovascular monitoring is intended to avoid. The 1 s operating point is therefore representative of continuous-monitoring schedules rather than an arbitrary single configuration. A 1 s cadence is also consistent with established practice in real-time ECG telemetry. Marouf et al. \cite{marouf2017multi} buffer ECG streams in 1 s windows for filtering, visualization, heart-rate extraction, and signal-quality feedback. Kakria et al. \cite{kakria2015real} similarly transmit cardiac data from a wearable sensor to a smartphone at 1 Hz. Our 1 s budget therefore follows established precedent.

\subsection{Results}
\label{subsect:effresults}
\begin{table*}[width=\textwidth,cols=7,pos=h]
	\caption{Efficiency Results of Our CNN Model on the MCXN-947}
	\label{table:ResultsExtended}
	
	\begin{tabular*}{\tblwidth}{@{}l@{\extracolsep{\fill}}cccccc@{}}
		\toprule
		\multicolumn{1}{c}{\multirow{2}{*}{Model}} &
		\multicolumn{3}{c}{CPU: Arm Cortex-M33} &
		\multicolumn{3}{c}{NPU: eIQ Neutron} \\
		\cmidrule(lr){2-4}\cmidrule(lr){5-7}
		&
		\begin{tabular}[c]{@{}c@{}}Memory\\ Usage\\ {[}Bytes{]}\end{tabular} &
		\begin{tabular}[c]{@{}c@{}}Inference\\ Time\\ {[}ms{]}\end{tabular} &
		\begin{tabular}[c]{@{}c@{}}Energy\\ per Inference\\ {[}mJ{]}\end{tabular} &
		\begin{tabular}[c]{@{}c@{}}Memory\\ Usage\\ {[}Bytes{]}\end{tabular} &
		\begin{tabular}[c]{@{}c@{}}Inference\\ Time\\ {[}ms{]}\end{tabular} &
		\begin{tabular}[c]{@{}c@{}}Energy\\ per Inference\\ {[}mJ{]}\end{tabular} \\
		\midrule
		
		CNN 6000 &
		45152 &
		\begin{tabular}[c]{@{}c@{}}103.06\\ $\pm$ 0.608\end{tabular} &
		\begin{tabular}[c]{@{}c@{}}0.750\\ $\pm$ 0.009\end{tabular} &
		27968 &
		\begin{tabular}[c]{@{}c@{}}37.82\\ $\pm$ 0.603\end{tabular} &
		\begin{tabular}[c]{@{}c@{}}0.204\\ $\pm$ 0.005\end{tabular} \\
		\midrule
		
		CNN 1024 &
		43904 &
		\begin{tabular}[c]{@{}c@{}}83.13\\ $\pm$ 0.138\end{tabular} &
		\begin{tabular}[c]{@{}c@{}}0.642\\ $\pm$ 0.004\end{tabular} &
		26752 &
		\begin{tabular}[c]{@{}c@{}}22.87\\ $\pm$ 0.135\end{tabular} &
		\begin{tabular}[c]{@{}c@{}}0.115\\ $\pm$ 0.002\end{tabular} \\
		\midrule
		
		CNN 512 &
		43776 &
		\begin{tabular}[c]{@{}c@{}}81.28\\ $\pm$ 0.085\end{tabular} &
		\begin{tabular}[c]{@{}c@{}}0.636\\ $\pm$ 0.005\end{tabular} &
		26624 &
		\begin{tabular}[c]{@{}c@{}}21.49\\ $\pm$ 0.094\end{tabular} &
		\begin{tabular}[c]{@{}c@{}}0.105\\ $\pm$ 0.002\end{tabular} \\
		\midrule
		
		CNN 256 &
		43712 &
		\begin{tabular}[c]{@{}c@{}}80.17\\ $\pm$ 0.105\end{tabular} &
		\begin{tabular}[c]{@{}c@{}}0.606\\ $\pm$ 0.005\end{tabular} &
		26560 &
		\begin{tabular}[c]{@{}c@{}}20.58\\ $\pm$ 0.067\end{tabular} &
		\textbf{\begin{tabular}[c]{@{}c@{}}0.086\\ $\pm$ 0.007\end{tabular}} \\
		\midrule
		
		CNN 128 &
		43680 &
		\begin{tabular}[c]{@{}c@{}}79.76\\ $\pm$ 0.088\end{tabular} &
		\begin{tabular}[c]{@{}c@{}}0.617\\ $\pm$ 0.009\end{tabular} &
		26496 &
		\begin{tabular}[c]{@{}c@{}}20.36\\ $\pm$ 0.089\end{tabular} &
		\begin{tabular}[c]{@{}c@{}}0.090\\ $\pm$ 0.002\end{tabular} \\
		\midrule
		
		CNN 64 &
		43680 &
		\textbf{\begin{tabular}[c]{@{}c@{}}79.59\\ $\pm$ 0.109\end{tabular}} &
		\textbf{\begin{tabular}[c]{@{}c@{}}0.604\\ $\pm$ 0.008\end{tabular}} &
		26496 &
		\textbf{\begin{tabular}[c]{@{}c@{}}20.17\\ $\pm$ 0.112\end{tabular}} &
		\begin{tabular}[c]{@{}c@{}}0.091\\ $\pm$ 0.004\end{tabular} \\
		
		\bottomrule
	\end{tabular*}
\end{table*}

Table~\ref{table:ResultsExtended} summarizes the impact of input signal length and processor choice. Across all signal lengths, inference on the NPU is markedly more efficient than on the Cortex-M33. For our chosen CNN 256 model, the inference time on the NPU is 20.58 ms compared with 80.17 ms on the Cortex-M33. This translates into 0.086 mJ per inference on the NPU versus 0.606 mJ on the Cortex-M33. Therefore, inference on the NPU is approximately 3.9 times as fast as inference on the CPU and requires approximately one-seventh of the energy.

The input signal length itself drives a substantial efficiency change. Reducing the per-modality input length from 6000 to 256 samples lowers the NPU energy per inference from 0.204 mJ to 0.086 mJ (\(\sim\)2.4x reduction) and the inference time from 37.82 ms to 20.58 ms. Combined with the near-best accuracy of CNN 256 (Table~\ref{table:Results1}), this validates aggressive downsampling as a design lever for edge deployment of multi-modal cardiac classifiers.

\begin{figure}
	\centering
	\includegraphics[width=0.62\columnwidth]{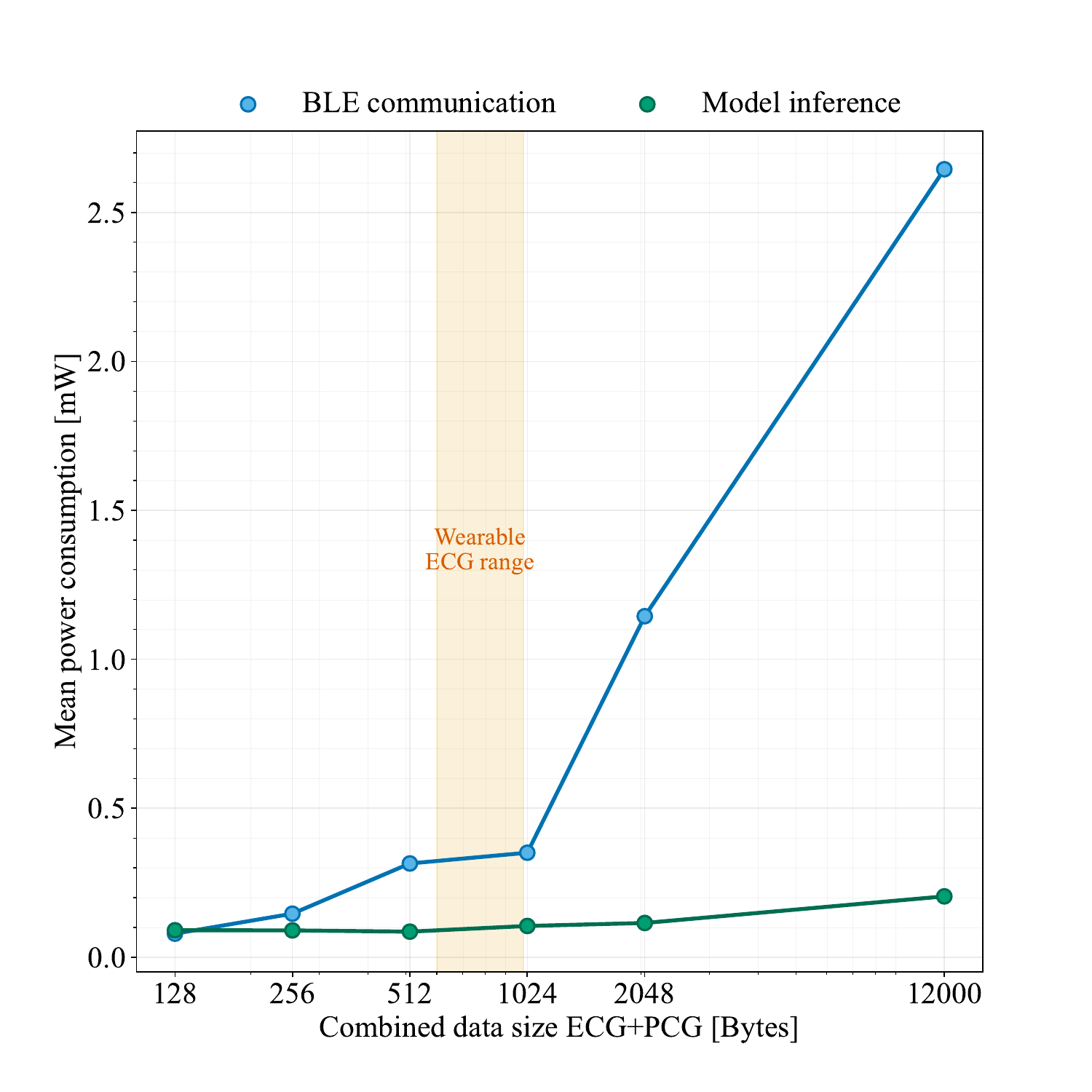}
	\caption{Mean power consumption of on-device neural-network inference (NPU) versus BLE communication, both operating at 1 Hz (one inference or one payload transmission per second). The two curves share a per-second data-volume axis. For BLE, this axis represents the transmitted payload. For inference, it represents the combined 8-bit ECG and PCG input size. Across the evaluated CNN variants, this input size ranges from 128 B for CNN 64 to 12000 B for CNN 6000. This shared axis places the two costs on the same scale. The shaded band marks realistic one-second wearable-ECG stream volumes (Section~\ref{sect:eff-A}). At this 1\,Hz cadence the mean power in mW equals the energy per second in mJ. The 12000 B/s BLE measurement uses a 15 ms connection interval to maintain stable sustained transmission. The nominal interval obtained from Eq. \eqref{eq:ci} is 20 ms.}
	\label{fig:PowerConsFig}
\end{figure}

Fig.~\ref{fig:PowerConsFig} reports the mean power consumption of model inference against BLE communication across the swept payload sizes. The average power during inference stays low across the swept range (0.086--0.204 mW, minimum at CNN 256), far less sensitive to data size than BLE. BLE communication, by contrast, increases strongly with payload size. At the smallest payload of 128 Bytes it is comparable to NPU inference, but it rises sharply for larger payloads and even exceeds model inference on the Cortex-M33 at a payload of 2048 Bytes, reaching 1.145 mW. The highlighted range in Fig. \ref{fig:PowerConsFig} marks the one-second wearable-ECG payloads introduced in Section \ref{sect:eff-A}. For this representative range, continuous streaming costs several times more than local model inference.

\subsection{Communication-and-Compute Energy Budget for Continuous Monitoring}
\label{subsect:budget}
The measurements above compare a single inference with a single transmission. We now extrapolate these measurements to continuous operation to address the design question introduced in Section~\ref{sec:introduction}. Is it more energy efficient to perform inference on the patch and transmit on demand, or to stream the signal continuously? Both operating strategies require the same always-on sensing, whose energy is outside the scope of this incremental comparison. We therefore compare the workload-specific inference and communication increments defined in Section~\ref{sect:eff-A}. The projection therefore concerns the communication and compute energy, rather than absolute device lifetime, which also includes the always-on sensing front-end that is common to any continuous-monitoring scheme. The value of moving inference on-device is precisely that it removes the dominant radio cost.

\begin{table}[width=\columnwidth,cols=5,pos=h]
	\caption{Projected daily energy budget for continuous monitoring, obtained by scaling the measured mean powers to one day.}
	\label{table:budget}
	\setlength{\tabcolsep}{3pt}
	\begin{tabular*}{\tblwidth}{@{}l@{\extracolsep{\fill}}cccc@{}}
		\toprule
		Strategy & \begin{tabular}[c]{@{}c@{}}Payload\\ per s\end{tabular} & \begin{tabular}[c]{@{}c@{}}Mean\\ power\\ {[}mW{]}\end{tabular} & \begin{tabular}[c]{@{}c@{}}Energy\\ per day\\ {[}J{]}\end{tabular} & Rel. \\ \midrule
		On-device inference (CNN 256, NPU) & 512 B   & 0.086 & 7.4   & 1$\times$    \\
		Continuous BLE stream               & 512 B   & 0.31  & 26.8  & 3.6$\times$  \\
		Continuous BLE stream               & 2048 B  & 1.145 & 98.9  & 13$\times$   \\
		Continuous BLE stream               & 12000 B & 2.64  & 228   & 31$\times$   \\ \bottomrule
	\end{tabular*}
	\\[2pt]{\footnotesize The inference row's 512 B is the model's input size, not a transmitted payload; it is placed on the same byte axis only to put compute and communication on a common per-byte scale (cf.\ Fig.~\ref{fig:PowerConsFig}).}
\end{table}

Table~\ref{table:budget} scales the measured mean powers to a per-day budget. Continuous on-device inference with the deployed CNN 256 draws 0.086 mW, about 7.4 J per day. A 16-bit, single-lead wearable ECG sampled at 300-500 Hz produces approximately 0.6–1.0 KB/s. At the original 2 kHz acquisition rate, the raw 16-bit dual-modality signal produces approximately 8 KB/s. The realistic streaming regime therefore lies toward the upper end of the tested data-rate range in Table~\ref{table:budget}, where power ranges from 1.145 mW at 2048 B/s to 2.64 mW at 12000 B/s. These values are respectively 13 and 31 times the measured power of on-device inference.

On-device processing replaces this continuous radio cost with uploads triggered only by a detected anomaly. Two properties keep the upload term small enough even in the worst case. First, the size of each upload is bounded. A context of 30 s (N=10 windows) corresponds to approximately 9–15 KB, depending on the sampling rate. This range surrounds the largest payload tested: transmitting 12 KB required 2.64 mJ. We therefore use 2.64 mJ as a representative estimate of the energy required for an upload in this size range. Second, the alarm rate, not the clinical-event rate, governs uploads.

To obtain a deliberately pessimistic stress-test estimate, we assume one classification decision per second and apply the measured false positive rate of $1-0.8966=0.1034$ directly to all $86\,400$ daily decisions. We further assume that every false positive decision triggers a separate upload, without suppressing or merging consecutive alarms. Under these assumptions, the model would produce approximately $86\,400 \cdot 0.1034 \approx 8934$ uploads per day. This estimate is not intended as a prediction of the alarm rate in a deployed system. In practice, the rolling majority vote described in Section~\ref{sect:method-D}, alarm aggregation, refractory periods, and application-specific decision thresholds would substantially reduce the number of distinct uploads. At 2.64 mJ per upload, the stress-test scenario corresponds to approximately 23.6 J of upload energy per day. Together with the 7.4 J/day inference cost, this yields an on-device communication and compute budget of approximately 31 J/day, which remains below the estimated raw-streaming cost of 99–228 J/day. At lower system-level alarm rates, as expected after temporal aggregation and alarm suppression, the on-device energy budget would be dominated by the 7.4 J/day inference term. A deployed patch would additionally consume energy for the lightweight preprocessing described in Section~\ref{sect:method-B}. These costs are expected to remain small relative to the continuous-streaming radio cost. This quantifies this work's central claim and delimits the realistic regime in which on-device inference is the more energy-efficient choice.

\section{Discussion}
\label{sect:disc}
This work provides a proof of concept demonstrating the feasibility of running an end-to-end, multi-modal DL model on resource-constrained medical edge devices for cardiovascular monitoring. We present quantitative evidence that local processing on an NPU with a lightweight CNN can be more energy efficient than constant data transmission over BLE. We also show that competitive classification accuracy can be retained despite an aggressive reduction in model size and input signal resolution. From a pervasive and mobile computing perspective, this shifts the system's energy bottleneck from the radio to computation and shows that an always-on wearable device can be made largely self-sufficient by keeping inference and the raw signal on the device.

\subsection{Model Performance and Efficiency}
\label{subsec:VI-A}
The signal-length sweep in Tables~\ref{table:SigLengths} and \ref{table:ResultsExtended} highlights the central trade-off between accuracy and efficiency. Using the full 2 kHz signal yields a marginal accuracy benefit but a substantial energy penalty. CNN 256 provides a favorable accuracy–energy operating point. It achieves the lowest measured NPU energy per inference in the sweep, approximately 2.4 times lower than CNN 6000, while its accuracy remains within the run-to-run variation observed for the longer inputs. Although the optimal trade-off point will depend on the deployment context, including the acceptable false-positive rate, target battery life, and recording duty cycle, CNN 256 is a defensible default for continuous wearable-patch monitoring.

Table~\ref{table:Quant} shows that 8-bit quantization-aware training reduces accuracy from 0.9748 to 0.9055, with corresponding decreases in sensitivity, specificity, and AUC. This is a meaningful degradation, particularly because the lower specificity increases the expected number of false alarms and therefore the potential communication overhead in continuous monitoring. The result nevertheless demonstrates the feasibility of executing the complete integer model on the target NPU. It should not be interpreted as evidence that the current quantized model is ready for clinical deployment. Further optimization and validation are required before practical use. The relatively large degradation suggests that the current architecture and training procedure are sensitive to 8-bit quantization. One possible explanation is the model’s small parameter budget, which may provide less redundancy for absorbing quantization error. Promising directions include quantization-aware architecture search, improved quantization-aware training, and calibration using representative target-domain data.

Table~\ref{table:ResultsNAS} shows that, at a fixed parameter budget around 16 K, our CNN clearly outperforms the MLP, LSTM, and transformer baselines. We attribute this to two factors. First, the anomalies of interest (e.g., mitral valve prolapse) typically manifest as morphological changes within a single cardiac cycle \cite{bhutto1992electrocardiographic, birnbaum2014ecg}. CNNs excel at detecting such local patterns. The longer-range temporal correlations that LSTMs are good at appear less critical here. Second, transformers are known to require more data than 405 recordings to train effectively in the absence of strong inductive biases.

Table~\ref{table:Results2} supports the hypothesis that the electrical (ECG) and mechanical (PCG) signals are complementary. Simply combining them improves accuracy only marginally over the strongest unimodal baseline (ECG Only), but the affine transform layer adds a further 2.4 percentage points by letting the network re-weight the two modalities. Even very small DL models can therefore exploit multi-modal data effectively, which is encouraging for tightly constrained edge deployments.

Fig.~\ref{fig:PowerConsFig} illustrates the conditions under which on-device classification is more energy efficient than continuous BLE streaming. Even at the smallest BLE payload tested, model inference on the NPU is competitive with constant streaming, and at any payload above ~256 Bytes the NPU inference is substantially cheaper. Streaming energy increases with data volume, whereas inference energy remains approximately fixed for a given model. We therefore expect the qualitative trade-off to hold across comparable edge platforms, although the crossover payload will depend on the radio and accelerator. By transmitting only when an anomaly is detected, the overall system energy is dominated by inference rather than radio activity, leaving sufficient energy margin for occasional event-triggered transmissions. The memory overhead of buffering the N most recent windows is small since it requires about 256 bytes per modality per window for the 8-bit quantized model due to the heavy downsampling.

\subsection{Limitations and Future Work}
\label{subsec:VI-B}
A central challenge for continuous monitoring with wearable patches is robustness to motion artifacts and ambient noise. The pipeline presented here does not include explicit denoising, because the primary goal of this paper is to establish the baseline efficiency of on-device ML. For real-world deployment, the pipeline must be extended to account for motion artifacts and ambient noise. Several potential approaches are compatible with the efficiency requirements reported above.

The first is lightweight upstream signal-quality assessment. Standard frequency-domain SQI checks are expected to consume substantially less energy than the CNN. Windows that fail SQI can be discarded before inference, which suppresses artifact-driven false positives without retraining. The second is the rolling majority vote described in Section~\ref{sect:method-D}. Increasing N inherently smooths over isolated, artifact-corrupted windows at negligible memory cost. The third is data augmentation during training with synthetic or real motion artifacts and ambient noise. This changes only the training pipeline, not the inference cost. Characterizing the resulting trade-off between robustness and efficiency on real-world wearable recordings is an important direction for future work.

We trained and evaluated the model exclusively on the training-a subset of the PhysioNet 2016 Challenge. To our knowledge, this is the only publicly available dataset containing synchronized ECG and PCG recordings with cardiovascular disease labels. The other subsets of the 2016 challenge and the more recent PhysioNet 2022 challenge are PCG-only and therefore cannot be used to validate the multi-modal pipeline as a whole. This restricts the dataset diversity and is a real risk to generalization. In line with the SAGER (Sex and Gender Equity in Research) guidelines, we further note that the training-a subset provides no sex or gender metadata for its recordings. A sex- and gender-disaggregated analysis was therefore not possible. Consequently, the generalizability of the results across sexes and genders remains unknown.

To partially mitigate the resulting overfitting risk, we used 10-times repeated 5-fold CV. The consistently low standard deviations across all metrics in Tables~\ref{table:Results1}–\ref{table:Quant} indicate that the model is stable across partitionings. Nevertheless, validation on larger, more diverse, fully annotated multi-modal datasets is needed before the model can be considered clinically robust. Moreover, as discussed in Section~\ref{subsect:eval}, training-a does not provide patient-level identifiers. Consequently, recordings from the same patient may be assigned to different folds, potentially inflating the reported performance. Recording-level splitting eliminates the dominant leakage path (windows from the same recording), but the residual inter-recording correlation cannot be ruled out from this dataset alone.

A further limitation concerns the signal-acquisition domain. The training-a recordings were captured with a clinical electronic stethoscope and a single-lead ECG during supervised in-home and hospital visits, whereas our target application is an ambulatory adhesive patch. Wearable-patch signals differ from the challenge data in electrode and microphone coupling, motion-artifact prevalence, and baseline noise. Accordingly, the reported accuracy should be interpreted as a feasibility result rather than as validated wearable-patch performance. Bridging this acquisition-domain gap requires evaluation on data recorded with the target patch hardware and is an important direction for future work.

The \(\sim\)7 percentage point accuracy drop introduced by 8-bit quantization (Table~\ref{table:Quant}) leaves a meaningful gap between the floating-point accuracy reported in Tables~\ref{table:Results1}–\ref{table:Results2} and the accuracy that would be observed in a deployed integer-only pipeline. Closing it through quantization-aware NAS or post-training calibration is a clear next step.

\section{Conclusion}
\label{sect:conc}
We developed a lightweight CNN for classifying synchronized ECG and PCG recordings. At floating-point precision, the model achieves accuracy competitive with the best reported results while requiring approximately three orders of magnitude fewer parameters and FLOPs than the leading architectures.

Hardware measurements on the MCXN947 EVK and complementary BLE measurements on the NHS52S04 EVK show that NPU-based local inference requires less energy than continuous data transmission across a wide range of payload sizes. These results demonstrate the potential of on-device DL for resource-constrained medical edge devices.

Future work will incorporate auxiliary signals, such as accelerometer data, to improve contextual awareness and provide references for motion artifacts. We will also investigate the reliability and robustness of the complete pipeline using the approaches outlined in Section~\ref{subsec:VI-B}. More broadly, our results provide concrete guidance for edge–cloud partitioning in energy- and privacy-constrained mobile-health systems.

\appendix
\section{Deep Learning Architectures}
\label{DLArch}
The MLP model consists of four linear layers with ReLU activations and a final linear layer that maps to the output. The output dimensions of the intermediate layers are 37, 35, 55 and 54. The LSTM model consists of one LSTM layer with a hidden unit size of 61 followed by a final linear layer that maps the output features of the last time step to the output. The transformer model consists of one convolutional patch-extraction layer (kernel size 14, stride 3, 28 output channels), followed by two transformer encoder blocks. Each encoder block has 28 features and a token-wise feed-forward width of 77. After the encoder blocks, an average pooling along the time dimension is followed by a final linear layer that maps to the output.

\printcredits

\section*{Declaration of competing interest}
Four of the authors---Mustafa Fuad Rifet Ibrahim, Tunc Alkanat, Felix Manthey, and Maurice Meijer---are employed by NXP Semiconductors, and this study evaluates NXP hardware, namely the MCXN947 microcontroller with its integrated eIQ Neutron NPU and the NHS52S04 Bluetooth Low Energy SoC. Alexander Schlaefer (Hamburg University of Technology) and Peer Stelldinger (Hamburg University of Applied Sciences) declare no competing interests. To limit the influence of this interest on the comparison, the energy analysis relies on a vendor-neutral incremental energy methodology that subtracts a same-rail idle baseline from both the inference and the transmission path, and all measurements were obtained with standard, commercially available evaluation kits.

\section*{Funding}
Funding: This work was supported by the Bundesministerium für Forschung, Technologie und Raumfahrt (BMFTR) under grant number 16KISK221 ("Holistische Entwicklung leistungsfähiger 6G-Vernetzung für verteilte medizintechnische Systeme (6G-Health)").

\section*{Ethics statement}
This study is a secondary analysis of the publicly available, fully de-identified PhysioNet/Computing in Cardiology Challenge 2016 database. No new data were collected from human participants by the authors, and the database provides no identifiers linking recordings to individuals. Ethical approval and informed consent for the original data collection were the responsibility of the contributing institutions. No additional institutional review board approval was required for this retrospective analysis of anonymized data.

\section*{Data availability}
The classification experiments use openly available data from the PhysioNet/Computing in Cardiology Challenge 2016 database~\cite{challenge2016data} (subset \emph{training-a}), described by Liu et al.~\cite{liu2016open}. No new physiological data were collected from human participants.

\section*{Code availability}
The source code for model training and evaluation is available from the corresponding author on request during peer review and will be deposited in a public repository with an archival DOI upon publication. The embedded firmware used for the on-device energy benchmark depends on proprietary vendor tooling and cannot be released in a public repository. However, the released software together with the measurement protocol in Section~\ref{sect:eff-A} and the referenced evaluation kits and their SDKs are sufficient to reproduce the reported energy and latency results.

\section*{Declaration of generative AI and AI-assisted technologies in the manuscript preparation process}
During the preparation of this work the authors used Microsoft 365 Copilot for language editing and to improve the clarity and presentation of the manuscript. After using this tool/service, the authors reviewed and edited the content as needed and take full responsibility for the content of the published article.

\bibliographystyle{model1-num-names}

\bibliography{cas-refs}


\end{document}